# Implementing Online Reinforcement Learning
## *with*
## Clustering Neural Networks

**J. E. Smith**
**University of Wisconsin-Madison (Emeritus)**
**02/29/2024**

**Abstract**

An agent employing reinforcement learning takes inputs (state variables) from an environment and performs actions that affect the environment in order to achieve some objective. Rewards (positive or negative) guide the agent toward improved future actions. This paper builds on prior clustering neural network research by constructing an agent with biologically plausible neo-Hebbian three-factor synaptic learning rules, with a reward signal as the third factor (in addition to pre- and post-synaptic spikes). The classic cart-pole problem (balancing an inverted pendulum) is used as a running example throughout the exposition. Simulation results demonstrate the efficacy of the approach, and the proposed method may eventually serve as a low-level component of a more general method.

## 1. Introduction

In this paper, a Clustering Neural Network (C*l*NN) architecture for online reinforcement learning (RL) is proposed and studied via simulation. The RL paradigm employs biologically plausible neo-Hebbian three-factor learning rules as articulated by Gerstner et al. [4].

A basic RL system is illustrated in Figure 1. Given the *state* of the *environment*, an *agent* decides on an *action* in order to achieve an objective. When determining the action, the agent relies on synaptic weights which encode information learned from past experience. The agent's action then affects the environment. If it increases the chances of achieving the goal, a *reward* is signaled to the agent. Or, if the action reduces the chances of achieving the goal, a punishment (negative reward) is signaled. Not every action results in a reward or punishment.

In a running example used throughout this paper, the agent attempts to balance an inverted pendulum – the classic cart-pole problem. Success for a given time interval results in a reward. Failure (e.g., the pole falls) results in punishment.

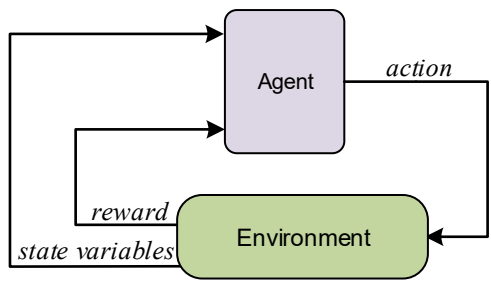

**Figure 1. An agent interacts with its environment in order to achieve an objective.**

## 1.1 Clustering Neural Networks

The neural networks described here are a specific type of spiking neural network distinguished in the following ways.

1) The presence (or absence) of individual spikes conveys information, rather than the rate at which spikes are emitted [6][12].

2) Synaptic weights are updated via a specialized form of spike timing dependent plasticity [1][7].

3) Neurons are modeled with active dendrites [5], and the basic computational function is performed by a dendritic segment. A single segment performs a function that is similar to that of a classical point neuron.

4) A dendrite performs unsupervised clustering and is an essential building block implemented by combining parallel segments with a winner-take-all (WTA) circuit [9][13].

5) Network operation is synchronized, with a single layer of segment/WTA functionality per synchronization cycle [1].

The architecture of C*l*NNs, as applied to online clustering and supervised classification, is covered in [11] – an essential prerequisite for this paper. C*l*NNs are given the additional capability of reinforcement learning by adding biologically plausible three factor learning rules.

## 1.2 Running Example: Cart-Pole Problem

The cart-pole problem serves as a running example for describing the RL architecture. The formulation given here uses Nagendra et al. [8] as a starting point.

The objective is to learn to balance a pole attached to a moving cart that sits on a track of fixed length (Figure 2). This is to be done by applying forces +F and -F to the cart.

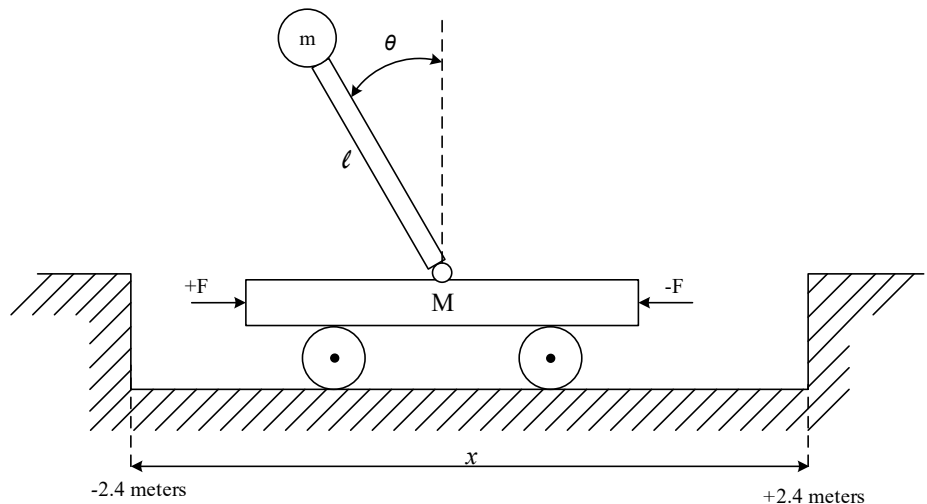

**Figure 2. Cart-Pole dimensions.**

Equations describing the system, taken from [8], are given below.

$$\ddot{\theta} = \{(M+m)g\sin\theta - \cos\theta[F+ml\dot{\theta}^2 \sin\theta]\} / \{(4/3)(M+m)l - ml\cos^2\theta\}$$

$$\ddot{x} = \{F+ml[\dot{\theta}^2\sin\theta - \ddot{\theta}\cos\theta]\} / (M+m)$$

Parameters for simulations follow; also taken from [8]

  $M$ = .711 kg        (1.6 lbs)
  $m$ = .209 kg        (.46 lbs)
  $g$ = 9.8 m/sec$^2$      (32 ft/sec$^2$)
  $F$ = ± 10 newtons     (2.25 lbs-force)
  $l$ = .326 meters      (1.1 ft)
  $\tau$ = .02 sec. time intervals

In the implementation described here, two state variables are used: the angle of the pole $\theta$ and the cart's velocity $dx/dt$. Failure occurs if the angle falls out of range (± 12 degrees) or the cart hits the end of the track.

## 2. RL Architecture

The model RL system is shown in Figure 3. Reinforcement Learning Neurons (RLNs) have proximal (P) and distal (D) inputs and perform inference in the same way as the neurons in [11]. The weight update function is extended to allow a reward (R) input which reflects the outcome of actions that occurred at some time in the past (e.g., several 10s of cycles). The environment (EV) is a part of the simulation system that models the cart-pole physics in 64-bit floating point and passes floating point state variables to encoding blocks.

The system here is constructed so that each of the possible actions (+F and -F) is implemented with an RLN. Each RLN groups input patterns appearing on the D inputs into clusters that invoke a specific action. During inference, both RLNs generate an output (A) whose value indicates the degree of similarity with other patterns associated with the RLN's assigned action. The WTA block selects the action with the stronger response (greater similarity). See [11] for details regarding the inference process.

The R inputs are used for weight updates as part of three-factor learning., and the P inputs are the inference outputs which act as enable signals for learning. Note that these particular neurons have only a single dendrite, hence a single P input line.

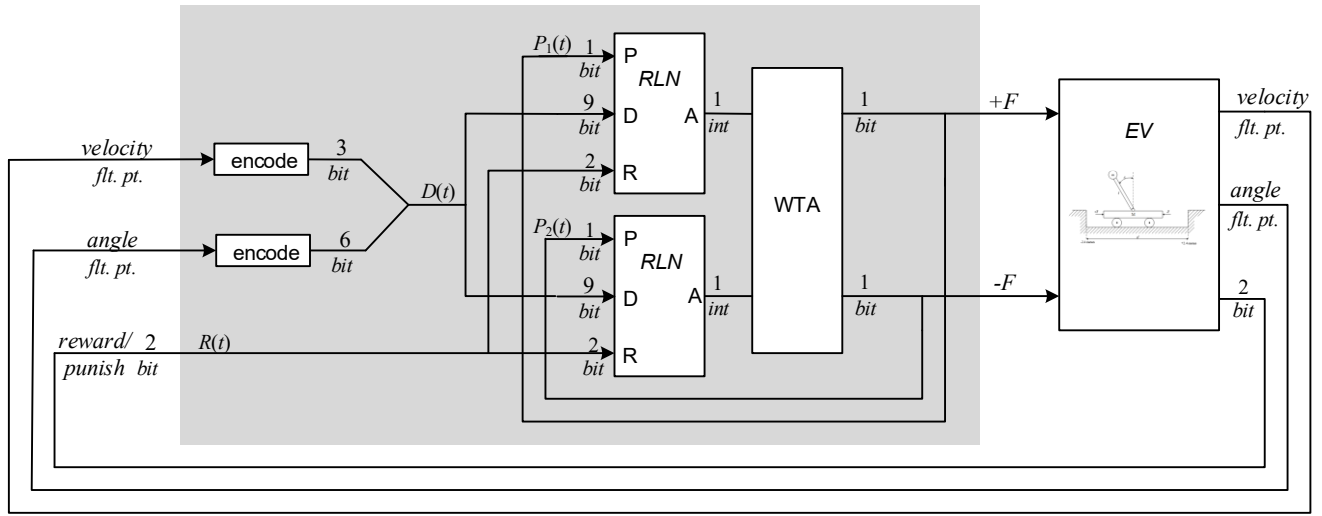

**Figure 3** Simulated cart-pole RL system. The schematic notation is described in the Appendix of [11]. The dimensions are for the two state variable system (cart velocity and pole angle) discussed later in Section 4. The spike trains $P(t)$, $D(t)$, and $R(t)$ tie together this schematic and schematics to follow.

## 2.1 Input Encoding

A simulated environment (EV) implements the cart-pole equations in floating point and produces state variables as output. The encoders reduce the incoming state variables to small discrete ranges, expressed as 1-hot bit vectors. For the problem at hand, the pole angle and cart velocity are encoded as shown in Table 1. To encode multiple state variables, multiple vectors are concatenated into a single larger vector.

**Table 1. Binary Encoding of State Variables**

|  | interval# | interval | encoding |
|---|---|---|---|
| angle θ | 1 | -12, -6 | 100000 |
|  | 2 | -6, -1 | 010000 |
|  | 3 | -1, 0 | 001000 |
|  | 4 | 0, 1 | 000100 |
|  | 5 | 1, 6 | 000010 |
|  | 6 | 6, 12 | 000001 |
| cart velocity d$x$/d$t$ | 1 | -$inf$, -5 | 100 |
|  | 2 | -5, +5 | 010 |
|  | 3 | 5, +$inf$ | 001 |

## 2.2 Reinforcement Learning Neurons

The schematic for an RLN is shown in Figure 4. Inference is the same as in [11]. What distinguishes RLNs is the weight update function described in the next subsection.

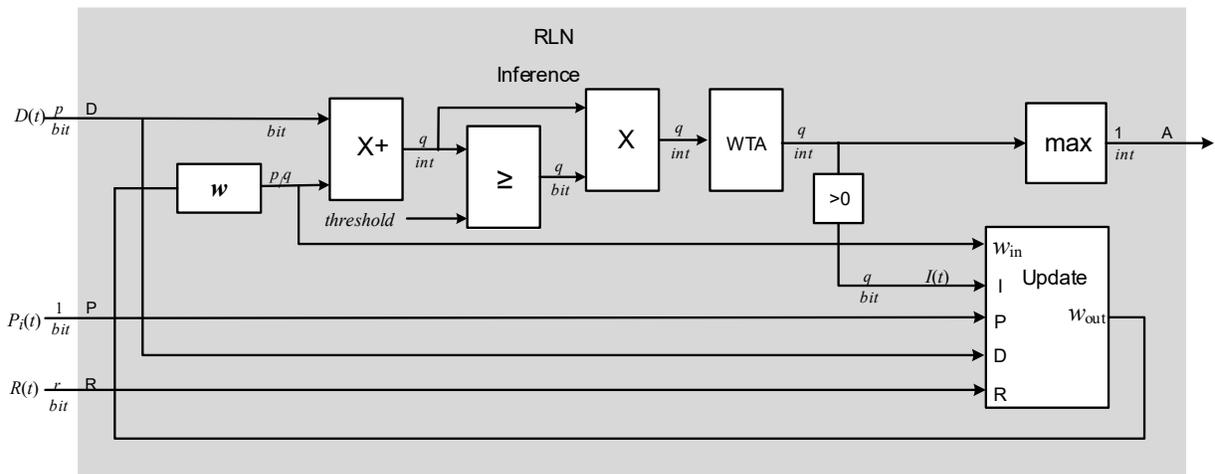

**Figure 4. Schematic for an RLN containing a single dendrite composed of $q$ segments. Inference is shown in detail. Note that the proximal input only applies to update, not inference. In effect, inference is enabled every cycle. Detail for the update block is in Figure 5 and Figure 6.**

## 2.3 Weight Update

Each RLN contains a single dendrite composed of segments that feed a WTA block associated with the dendrite. Say that during inference the most strongly stimulated segment $j$ is the dendrite's "winner". Then, if segment $j$ also wins at the network WTA (Figure 4), it is segment $j$ that ultimately causes an action to take place. Consequently, the sequence of winning segments, including $j$ in this case, form an *action path*. When there is a reward or punishment, credit assignment determines which of the prior actions along the action path should be rewarded (by potentiating associated synapses) or punished (by depressing associated synapses).

The neo-Hebbian three-factor learning rules as described by Gerstner et al. [4] lead to a promising biologically plausible credit assignment method. With this approach, the section of the action path to be updated begins at the *head* of the path (the current action) and extends back by a parameterized length until it reaches the *tail* which is the oldest previous action to be updated. In response to a reward, synaptic weights along the path are potentially updated. The amount of weight update is maximum at the head of the path and decays along the length of the path to the tail.

Observe that rather than incrementally propagating rewards and punishments over multiple cycles as in classic Q-learning [14], all the synaptic weight adjustments occur immediately when there is a reward or punishment.

With three-factor rules, the first two factors are the same as with conventional two-factor SDP [11]: a synapse's input spike and its post-synaptic neuron's output spike. The third factor is a broadcast reward signal $R$. $R$ is expressed as a 1-hot code, and at each time step there can be a positive reward, a negative reward, or no reward (actually 0-hot). For the cart-pole example, there is a single type of positive reward and a single type of negative reward; these are encoded as [1 0], [0 1], and [0 0], respectively.

In the cart-pole implementation positive rewards and negative rewards are handled differently. Negative rewards are implemented in accordance with error related negativity (ERN) [1]. However, applying negative rewards alone results in monotonic decreasing weights. To counteract this, positive rewards are included. For the cart-pole problem, there is a positive reward whenever the pole is successfully kept upright for intervals of a specified amount of time (number of steps).

### 2.3.1 Negative Rewards

Negative reward updates are implemented as shown in Figure 5 for synapse $i,j$. The three learning factors are the inputs $D_i$, $I_j$, and the broadcast reward $R$. For this network, $R = 0$ or -1. Just as with SDP, each synapse has its own local update logic, the only shared signal is the broadcast reward. When a negative reward is broadcast, the synapses of segments along the action path are subject to three-factor SDP update.

The implementation of synapse $i,j$ consists of the following parameters and state.

$\omega$: length of action path from head to tail that is subject to SDP updates.

$\pi$: maximum weight decrement (punishment).

$w^i_j$: weight – an up/down counter that saturates at 0 and $w_{max}$. Weight initialization is discussed in Section 3.2.

$c^i_j$: a decay counter that is set the maximum value of $\omega$ whenever there are spikes on $D_i$, $I_j$, and the proximal input $P$. At each subsequent step it counts down until it reaches 0.

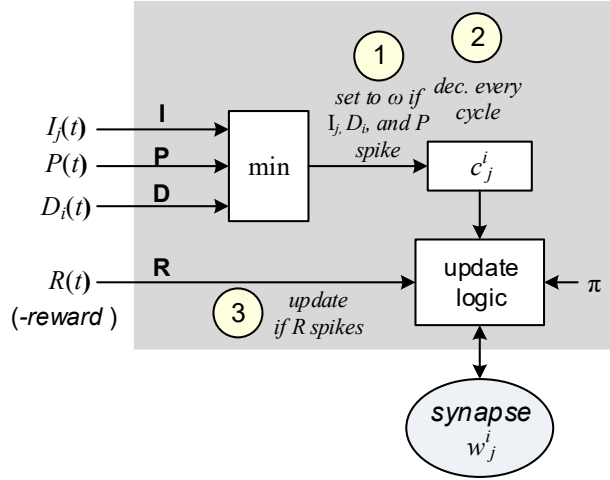

**Figure 5.** The RLN synapse implementation for negative rewards employs a neo-Hebbian three factor update rule. A decay counter is initialized to value ω when an input spike is followed by an output spike originating with the synapse's associated segment. The decay counter decrements every cycle until it reaches 0. When a negative reward is received, the synaptic weight is decreased by an amount that depends on the value of the decay counter and the maximum decrement π.

Whenever a negative reward is signaled, all synaptic weights with $c^i_j > 0$ are updated. Specifically:

$$\Delta w^i_j = -\pi \cdot c^i_j / \omega$$

Hence, the weight update values decay linearly from the most recently active synapse back to synapses up to $\omega$ steps in the past. Note that a given synapse is only decremented once, by a value that depends on the decay counter value for the synapse's most recent use. Also note that the decay function does not have to be linear. For example, one could use exponential decay, although it would likely require a more complex implementation.

### 2.3.2 Positive Rewards

Positive reward updates are implemented as shown in Figure 6. A positive reward is determined by a *success counter* (not shown in the figure). Starting at 0, in increments every successful step. When it reaches the value of $\sigma$, it signals a positive reward and resets to 0.

In response to the positive reward, synapses along the action path are updated. The operation is similar to negative rewards. The major difference is that when a reward is signaled, all the synapses belonging to a segment on the path are subject to update, not just the ones that received input spikes. Consequently, a flag is added to the update function to indicate whether or not the synapse received an input spike.

When a positive reward is broadcast, all the synapses belonging to all the segments in the success window $\sigma$ are subject to three-factor SDP update. The implementation of synapse $i, j$ consists of the following parameters and state.

    $\sigma$: size of the success window.
    $\rho^+$: the maximum weight increment; applied to synapses that receive an input spike.
    $\rho^-$: the maximum weight decrement; applied to synapses that do not receive an input spike.
    $w^i_j$: weight – an up/down counter that saturates at 0 and $w_{max}$. For initialization see Section 3.2.
    $c^i_j$: a decay counter that is set to value of $\sigma$ when there are spikes on $I_j$, $D_i$, and the proximal input P. At each subsequent step it counts down until it reaches 0.
    $e^i_j$: a binary flag that sets to 1 if input $D_i$ spikes. It is cleared to 0 if $D_i$ does not spike.

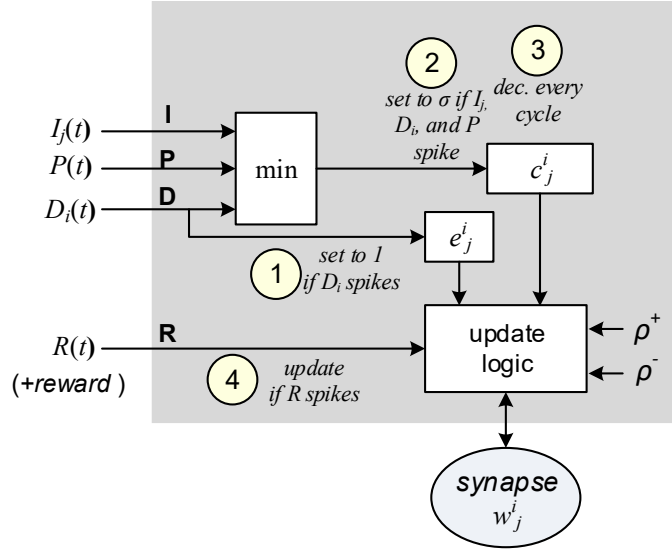

**Figure 6.** The RLN synapse implementation for positive rewards employs a neo-Hebbian three factor update rule. The *e* flag is set when there is an input spike, and the decay counter is initialized to $\sigma$ if the associated neuron produces an output spike. The decay counter decrements every cycle until it reaches 0. When a positive reward is received, the synaptic weight is updated, up or down, depending on whether the *e* flag is set and by an amount that depends on the value of the decay counter and the maximum increment or decrement ($\rho^+$ or $\rho^-$).

Whenever a positive reward is signaled, all synaptic weights with $c_j^i > 0$ are updated. Specifically:

$$\text{if } e_j^i = 1 \quad \Delta w_j^i = \rho^+ \cdot c_j^i / \sigma$$
$$\text{if } e_j^i = 0 \quad \Delta w_j^i = -\rho^- \cdot c_j^i / \sigma$$

Hence, the weight updates decay linearly from the most recently active synapse back to synapses $\sigma$ steps in the past. Parameters $\rho^+$ and $\rho^-$ correspond to the *capture* and *backoff* parameters used in[11] when forming clusters.

Note that the same update parameters apply to all neurons and synapses. They are established at the time the system is constructed, and they are not affected by the learning process.

## 3. Simulations: One State Variable (1 SV)

For the first set of simulations only one state variable is used: the pole angle. The pole angle is restricted to the range of ± 12°, and this range is discretized into 6 intervals and 1-hot encoded as shown in Table 1. When the pole angle exceeds the range, the trial is deemed a failure. The cart reaching the end of the track is also a failure. Hence, there are two causes of negative rewards, but both are handled in the same way by the 3-factor update mechanism.

### 3.1 Optimal

An optimal method is implemented as a standard for comparison. With only one state variable, if the pole leans left, then force should push the cart to the left (-F = 1), and if the pole leans right, force should be applied to push the cart to the right (+F = 1). In the absence of any other state information, this naive approach is also optimal. Accordingly, for the first set of simulations, weights were manually set to achieve the optimal strategy, see Figure 1.

|  | Angle | | | | | |
|---|---|---|---|---|---|---|
| -F | 1 | 2 | 3 | 4 | 5 | 6 |
| Segment 1 | 8 | 0 | 0 | 0 | 0 | 0 |
| Segment 2 | 0 | 8 | 0 | 0 | 0 | 0 |
| Segment 3 | 0 | 0 | 8 | 0 | 0 | 0 |

|  | Angle | | | | | |
|---|---|---|---|---|---|---|
| +F | 1 | 2 | 3 | 4 | 5 | 6 |
| Segment 1 | 0 | 0 | 0 | 0 | 0 | 8 |
| Segment 2 | 0 | 0 | 0 | 0 | 8 | 0 |
| Segment 3 | 0 | 0 | 0 | 8 | 0 | 0 |

**Figure 7. Optimal synaptic weights for -F and +F neurons.**

Using the optimal weights (so no training is needed), each of the 32 potential starting angles was simulated. The results are in Figure 8. For angles close to 0, the maximum of 10,000 steps is achieved. For larger angles, the number of successful steps varies from as low as 3,000 steps to as many as 9,000. The differences are dependent on the intervals used for angle encodings (Table 1). The average number of steps across all 32 trials is 6380.

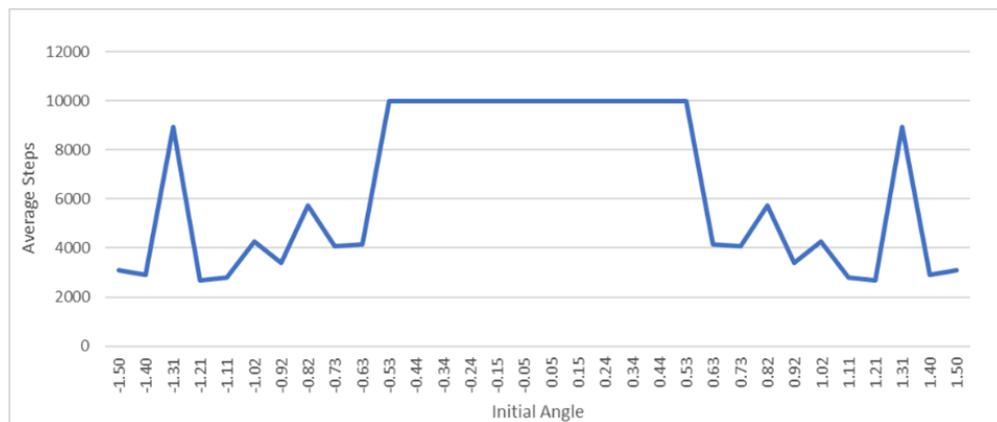

**Figure 8. Average steps for each of the 32 initial angles between -1.5 and +1.5 degrees, using optimal synaptic weights.**

For reference, if no force is applied and the pole is simply allowed to drop, it takes 275 steps for the pole to fall out of range. If a constant positive (or negative) force is applied and maintained, it takes 32 steps for the cart to go out of range.

## 3.2  3-Factor Simulations

The same methodology is used for both the single state variable simulations described in this section and for the two state variable simulations in the next section.

A single simulation run (episode) consists of a sequence of 512 trials. The network continually learns over the sequence of trials. For each trial the cart is initially placed in the center of the track ($x(0) = 0$), and the angle is initialized to one of 32 equally spaced values in the range $\pm 1.5°$. The entire sequence of initial angles consists of repeated random permutations of the 32 possibilities.

Weights are initialized to a base value $w_{base}$ plus a small pseudo-random value between 0 and $w_{range}$. That is the initial weight = $w_{base} + rand(0, w_{range})$. The rationale for adding a small random value is that weight updates may occur several cycles after the weight is accessed for inference. If weights are all initialized to the same constant, the initial spike patterns applied to the network often result in WTA ties. To break the tie, a random segment could be selected. If, before weights are updated, the same input pattern is applied there will again be a tie. In this case, it is desirable for WTA to select the same segment as for the first application of the pattern. Adding a small pseudo-random value to the initial weights significantly reduces the number of ties so segment selection retains its random character, and the same input pattern consistently maps to the same segment.

After initialization, each trial is executed until the pole falls out of range $\pm 12º$, $x$ exceeds $\pm 2.4$ meters, or 10,000 successful steps have been performed (200 seconds of simulated time).

The performance metric for a given episode is the average successful steps over all trials, including the very first trials where most of the learning takes place.

It was found that the seed for generating pseudo-random initial angles has a significant effect on the results. To avoid inadvertent cherry-picking, 32 different random seeds were used, and simulation results for all 32 episodes are given in plotted results to follow.

For these simulations, there are 3 segments per neuron, $w_{max} = 8$, $w_{base} = 9/2$, and $w_{range} = 1/128$.

Note that although weights may contain fractional values, in simulations, the ceiling function is applied to weights before they are used for inference, consequently, inference is an entirely small integer operation.

After a series of preliminary simulations to sweep the parameter space under manual direction, the simulation parameters in Table 2 were chosen. In general, performance is relatively insensitive to the specified parameter values.

Table 2. Parameters for Simulations

| parameter | value | description |
|---|---|---|
| $\sigma$ | 256 | Success window size |
| $\omega$ | 256 | Punishment update window size |
| $\rho_0^+$ | 1/128 | Maximum reward capture amount |
| $\rho_0^-$ | 7/1024 | Maximum reward backoff amount |
| $\pi$ | 8/1024 | Maximum punishment backoff amount |

Results for the 32 episodes using different random number seeds are plotted in Figure 9, along with the optimal result. The actual implementation performs nearly as well as the optimal case. The small difference is due to the initial trials that begin with an untrained network; the optimal network begins in a fully trained state.

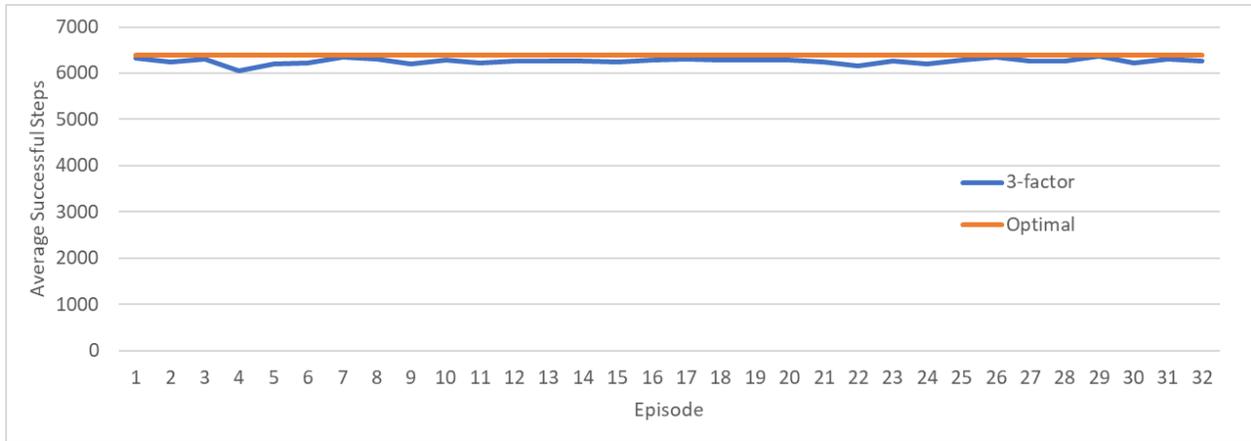

**Figure 9. Average number of successful steps over 512 trials using 32 random number generator seeds. For comparison, the averages for optimal weights are also given.**

For the dimensions and parameters used here, if there is a failure, it is typically because the cart hits the end of the track. The common scenario is that the pole is kept within range for long periods of time, but there is a slow drift in cart displacement (Figure 10). Eventually, the cart drifts to the end of the track. Of course, this behavior is to be expected because displacement is not used as a state variable. In contrast, Figure 11 illustrates behavior for a typical section of a successful trial.

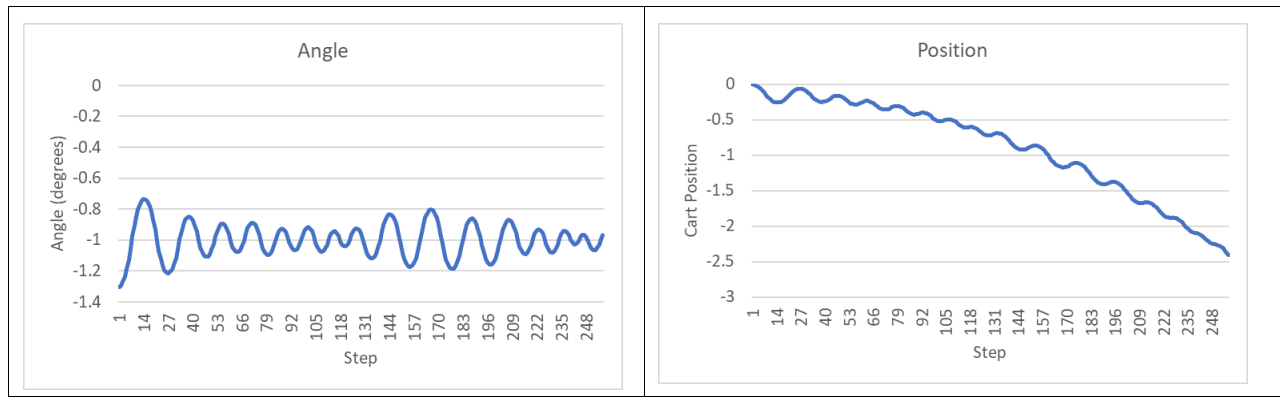

**Figure 10. Pole angle and cart position as a function of the step. Results for a typical failing simulation are shown. The cart gradually drifts to the left until it hits the end of the track. Meanwhile, the pole angle remains within valid range.**

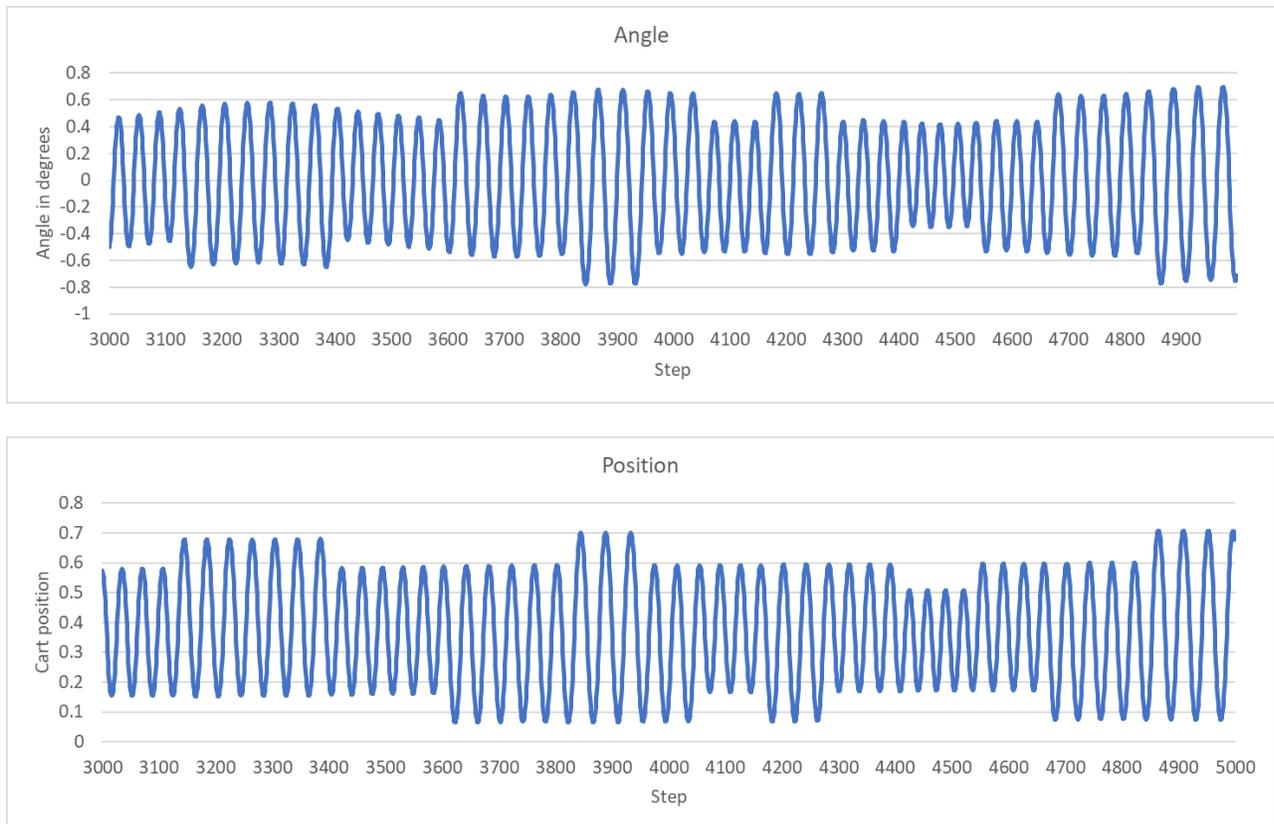

**Figure 11** Pole angle and cart position as a function of the step. A section of a successful simulation is shown.

### 4. Simulations: Two State Variables (2 SV)

Although the single state variable system just described seems to perform reasonably well, most trials fail because the cart eventually drifts to an end of the track (Figure 10). To improve performance, a second state variable, the cart velocity is added to allow the agent to adjust for the cart drift. See Table 1 for d$x$/d$t$ ranges and encoding.

### 4.1 Optimal

For the 2 SV system, there are some cases where it is better not to follow the optimal 1 SV method. In particular, if the pole is leaning left and the cart is strongly accelerating to the left, it is sometimes better to push the car to the right, +F = 1, in order to slow it down, rather than push it left to accommodate the angle of the pole. For 2 SVs, the optimal synaptic weight matrices are in Figure 12.

| | | Angle | | | | | | dx/dt | | |
|---|---|---|---|---|---|---|---|---|---|---|
| | **-F** | 1 | 2 | 3 | 4 | 5 | 6 | 1 | 2 | 3 |
| | | 8 | 0 | 0 | 0 | 0 | 0 | 0 | 8 | 0 |
| | | 8 | 0 | 0 | 0 | 0 | 0 | 0 | 0 | 8 |
| | | 0 | 8 | 0 | 0 | 0 | 0 | 0 | 8 | 0 |
| | | 0 | 8 | 0 | 0 | 0 | 0 | 0 | 0 | 8 |
| Segment | | 0 | 0 | 8 | 0 | 0 | 0 | 8 | 0 | 0 |
| | | 0 | 0 | 8 | 0 | 0 | 0 | 0 | 0 | 8 |
| | | 0 | 0 | 0 | 8 | 0 | 0 | 0 | 8 | 0 |
| | | 0 | 0 | 0 | 0 | 8 | 0 | 0 | 0 | 8 |
| | | 0 | 0 | 0 | 0 | 0 | 8 | 0 | 0 | 8 |

| | | Angle | | | | | | dx/dt | | |
|---|---|---|---|---|---|---|---|---|---|---|
| | **+F** | 1 | 2 | 3 | 4 | 5 | 6 | 1 | 2 | 3 |
| | | 0 | 0 | 0 | 0 | 0 | 8 | 0 | 8 | 0 |
| | | 0 | 0 | 0 | 0 | 0 | 8 | 8 | 0 | 0 |
| | | 0 | 0 | 0 | 0 | 8 | 0 | 0 | 8 | 0 |
| | | 0 | 0 | 0 | 0 | 8 | 0 | 8 | 0 | 0 |
| Segment | | 0 | 0 | 0 | 8 | 0 | 0 | 0 | 0 | 8 |
| | | 0 | 0 | 0 | 8 | 0 | 0 | 8 | 0 | 0 |
| | | 0 | 0 | 8 | 0 | 0 | 0 | 0 | 8 | 0 |
| | | 0 | 8 | 0 | 0 | 0 | 0 | 8 | 0 | 0 |
| | | 8 | 0 | 0 | 0 | 0 | 0 | 8 | 0 | 0 |

**Figure 12 Optimal synaptic weights for -F and +F neurons.**

The optimal performance for the 32 initial angles is plotted in Figure 13. The performance improves significantly over the 1 SV case, with nearly all initial angles leading to the maximum of 10,000 successful steps.

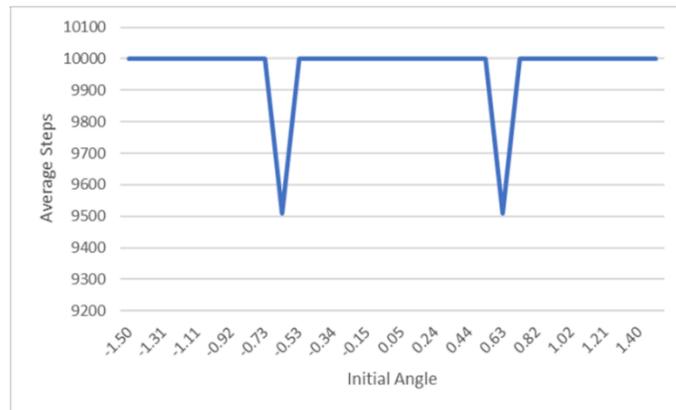

**Figure 13. Average steps for each of the 32 initial angles between -1.5 and +1.5 degrees, using optimal synaptic weights.**

### 4.2  3-Factor Simulations

All parameters are the same as for the single state variable case (Table 2 ). A total of 32 episodes with different random number seeds were simulated and results are plotted in Figure 14. Results are sorted according to the average number of successful steps from highest to lowest. For comparison, performance using optimal weights for both the one state variable system (1 SV) and two state variable system (2 SV) are plotted.

The results are mixed. There are three episodes where performance clearly benefits from the second state variable. Then there are another 12 episodes where the performance is about the same as for a single state variable, accounting for low-performing initial trials. However, about half the episodes have worse performance than for a single state variable. This is likely because there are multiple local optima, and adding a state variable increases the number of local optima, some of which are significantly worse than the global optimum. Figure 15 illustrates behavior for a typical section of a successful trial.

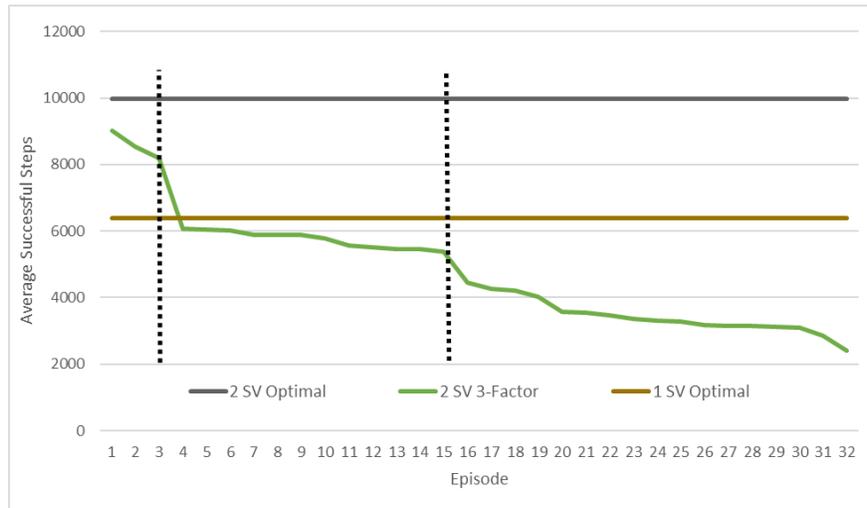

Figure 14. Average successful steps for 32 episodes with different pseudo-random seeds. Also plotted is performance for optimal systems having 1 and 2 state variables (1 SV and 2 SV)

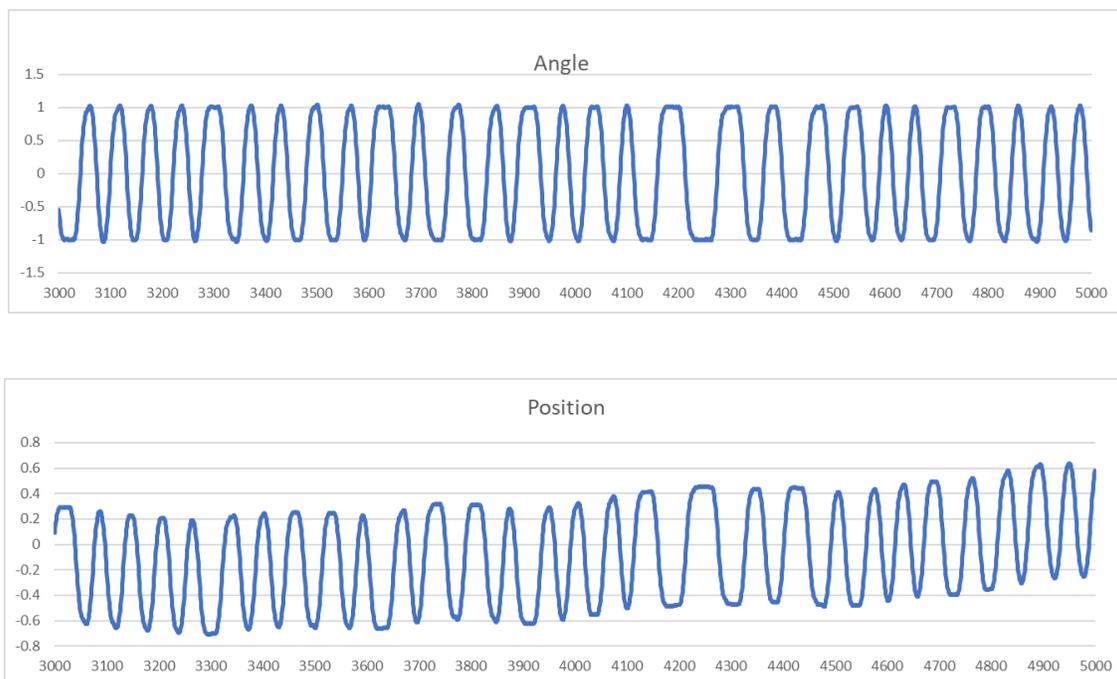

Figure 15. Pole angle and cart position as a function of the step for the two state variable system. A section of a successful simulation run is shown.

## 5. Concluding Remarks

The neo-Hebbian update rules articulated by Gerstner et al. [4] have been demonstrated to work in a biologically plausible spiking neuron system performing a reinforcement learning task. The proposed 3-factor learning method is biologically plausible, simple, and easy to understand. Overall, it does a good (but not optimal) job of balancing the pole.

Admittedly, the model has been demonstrated for a very basic application. Hence, it serves only as a starting point for more elaborate applications that will likely require extensions. Furthermore, at this point, it is clear that the RL approach given here is not a complete system. A more complete system would always achieve near-optimal performance for two state variables, which this system does only occasionally.

The challenge for developing a more complete system is the same as in many non-linear optimization problems: to somehow steer the system toward a global optimum and avoid becoming trapped in a local optimum. This is a version of the classical stability plasticity dilemma. One approach, yet to be explored, is to add some randomness to the convergence process with the goal of jolting the system out of a local optimum, in order for the system to settle into a more optimal state. Follow-on research will consider ways of doing so, say by periodically re-randomizing a small subset of the weights.

Finally, the reader may have noticed that for the cart-pole design given here, only trivial clustering is done. There are enough segments so that each unique input pattern can be assigned its own segment. For larger scale problems with finer granularities, similarity coding [10][11] can be used when the number of segments is limited.

## 6. References

[1] Abeles, Moshe. "Synfire chains." *Scholarpedia* 4, no. 7 (2009): 1441.

[2] Bi, Guo-qiang, and Mu-ming Poo. "Synaptic modifications in cultured hippocampal neurons: dependence on spike timing, synaptic strength, and postsynaptic cell type." *The Journal of neuroscience* 18, no. 24 (1998): 10464-10472.

[3] Gehring, William J., Brian Goss, Michael GH Coles, David E. Meyer, and Emanuel Donchin. "The error-related negativity." Perspectives on Psychological Science 13, no. 2 (2018): 200-204.

[4] Gerstner, Wulfram, Marco Lehmann, Vasiliki Liakoni, Dane Corneil, and Johanni Brea. "Eligibility traces and plasticity on behavioral time scales: experimental support of neohebbian three-factor learning rules." *Frontiers in neural circuits* 12 (2018): 53.

[5] Hawkins, Jeff and S. Ahmad, "Why Neurons Have Thousands of Synapses, A Theory of Sequence Memory in Neocortex," Numenta, Inc, Redwood City, California, Oct. 30, 2015.

[6] Kermany, Einat, et al., Tradeoffs and Constraints on Neural Representation in Networks of Cortical Neurons –*The Journal of Neuroscience* 30.28 (2010): 9588-9596.

[7] Markram, Henry, Joachim Lübke, Michael Frotscher, and Bert Sakmann. "Regulation of synaptic efficacy by coincidence of postsynaptic APs and EPSPs." *Science* 275, no. 5297 (1997): 213-215.

[8] Nagendra, Savinay, Nikhil Podila, Rashmi Ugarakhod, and Koshy George. "Comparison of reinforcement learning algorithms applied to the cart-pole problem." In *2017 International Conference on Advances in Computing, Communications and Informatics (ICACCI)*, pp. 26-32. IEEE, 2017.

[9] Natschläger, Thomas, and Berthold Ruf. "Spatial and temporal pattern analysis via spiking neurons." *Network: Computation in Neural Systems* 9, no. 3 (1998): 319-332.

[10] Purdy, Scott. "Encoding data for HTM systems." *arXiv preprint arXiv:1602.05925* (2016).

[11] Smith, James E. "Neuromorphic Online Clustering and Classification" *arXiv:2310.17797v1* (2023)

[12] Thorpe, Simon J., and Michel Imbert. "Biological constraints on connectionist modelling." *Connectionism in perspective* (1989): 63-92.

[13] Thorpe, Simon J. "Spike arrival times: A highly efficient coding scheme for neural networks." *Parallel processing in neural systems* (1990): 91-94.

[14] Watkins, Christopher JCH, and Peter Dayan. "Q-learning." *Machine learning* 8, no. 3 (1992): 279-292.